\documentclass[11pt]{article}

\usepackage[preprint]{acl}

\usepackage{times}
\usepackage{latexsym}

\usepackage[T1]{fontenc}

\usepackage[utf8]{inputenc}

\usepackage{microtype}

\usepackage{inconsolata}

\usepackage{graphicx}
\usepackage{booktabs}
\usepackage{amsmath}
\usepackage{graphicx}
\usepackage{wrapfig}
\usepackage{multirow}
\usepackage{multicol}
\usepackage{amssymb}
\usepackage{xcolor}
\usepackage[most]{tcolorbox}
\usepackage[table]{xcolor}
\definecolor{Green}{rgb}{0.0, 0.2, 1.0}
\definecolor{Red}{rgb}{0.8, 0.0, 0.0}
\newtcolorbox{takeaway}{
  colback=white,
  colframe=green!75!black,
  fonttitle=\bfseries,
  title=,
  boxrule=1pt,
  arc=4pt,
  auto outer arc,
  width=\linewidth,
  breakable,
}

\title{\textsc{Select2Reason}: Efficient Instruction-Tuning Data Selection for Long-CoT Reasoning}

\author{
 \textbf{ Cehao Yang\textsuperscript{1,2,3}}\footnotemark[1],
 \textbf{ Xueyuan Lin\textsuperscript{2,3,4}}\footnotemark[1],
 \textbf{ Xiaojun Wu\textsuperscript{1,2,3}},
 \textbf{ Chengjin Xu\textsuperscript{1,3}}\footnotemark[1],\\
 \textbf{ Xuhui Jiang\textsuperscript{1,3}},
 \textbf{ Honghao Liu\textsuperscript{2,3}},
 \textbf{ Hui Xiong\textsuperscript{2}}\footnotemark[2],
 \textbf{ Jian Guo\textsuperscript{3}}\footnotemark[2]
\\
 \textsuperscript{1}DataArc Tech Ltd.
 \\
 \textsuperscript{2}Hong Kong University of Science and Technology (Guangzhou)
\\
 \textsuperscript{3}IDEA Research, International Digital Economy Academy
 \\
 \textsuperscript{4}Hithink RoyalFlush Information Network Co., Ltd
 \\
 \texttt{\{cyang289,xlin058,wuxiaojun\}@connect.hkust-gz.edu.cn}, \texttt{xionghui@ust.hk}\\
\texttt{\{xuchengjin,jiangxuhui,liuhonghao,guojian\}@idea.edu.cn}
}

\begin{document}
\maketitle

\begin{abstract}
A practical approach to activate long chain-of-thoughts reasoning ability in large language models is to perform supervised fine-tuning on instruction datasets synthesized by strong large reasoning models, offering a cost-effective alternative to reinforcement learning. However, large-scale instruction sets incur significant training overhead, while effective strategies for automatic data selection still remain unexplored. We propose \textbf{\textsc{Select2Reason}}, a novel and efficient instruction-tuning data selection framework for long-CoT reasoning. From the perspective of emergence of rethinking behaviors like self-correction and backtracking, we investigate metrics that may determine the quality of long-CoT instructions. \textsc{Select2Reason} leverages a difficulty-aware reward model to estimate the learning value of questions and jointly incorporates a reasoning trace length-based heuristic through a weighted scheme for ranking to prioritize high-utility examples. Empirical results on OpenR1-Math-220k demonstrate that fine-tuning LLM on only 10\% of the data selected by our method achieves performance competitive with or superior to full-data tuning and open-source baseline across nine competition-level mathematical benchmarks and four broader reasoning tasks. Further experiments highlight the scalability in varying data size, efficiency during inference, and adaptability to other instruction pools of \textsc{Select2Reason} with minimal cost.
\end{abstract}

\section{Introduction}
\label{sec:intro}

Large reasoning models (LRMs)~\citep{o1, deepseek, gemini}, mark a significant leap in the complex reasoning abilities of large language models (LLMs). With the emergence of the long chain-of-thoughts (long-CoT) reasoning ability~\citep{cotsurvey}, these models exhibit human-like behaviors such as exploration, verification, reflection, and correction, allowing them to autonomously derive multi-branch and multi-step solutions via deliberate planning and backtracking~\citep{reasonsurvey, system2survey}.

\begin{figure}
    \centering
    \includegraphics[width=\linewidth]{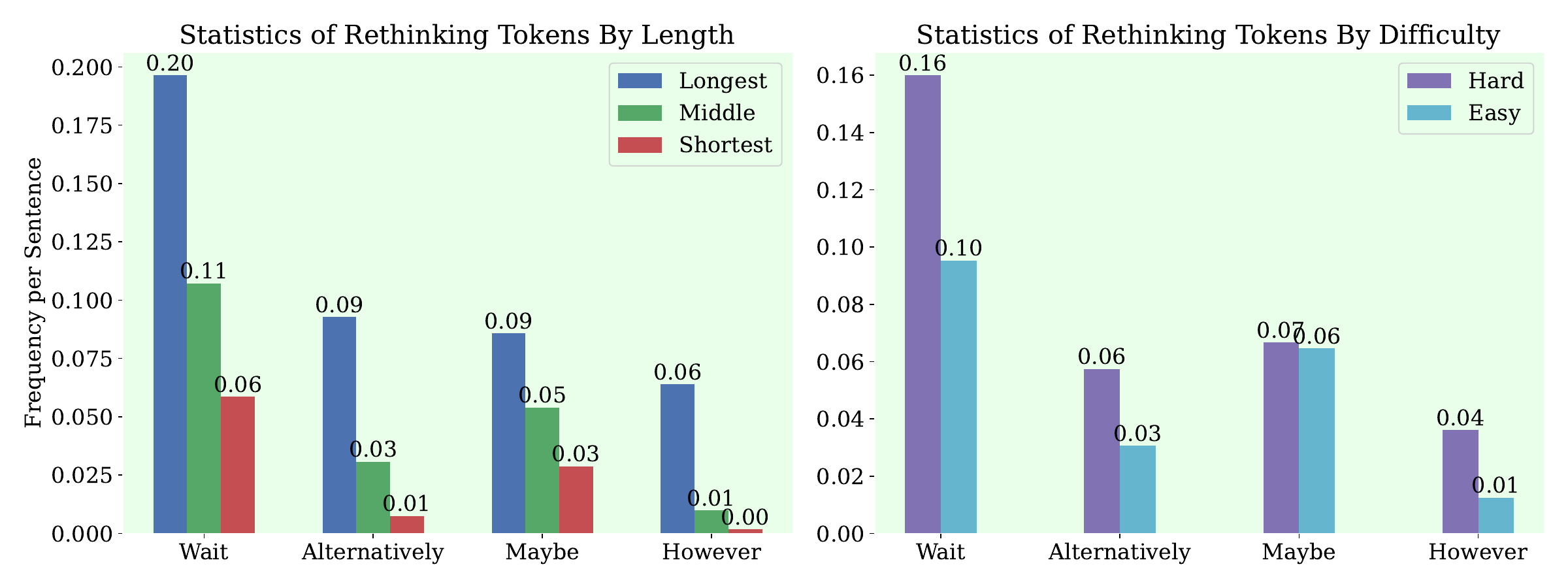}
    \caption{Statistics of rethinking tokens in reasoning trace. Longer reasoning traces exhibit a higher frequency of rethinking tokens in each step such as \textit{Wait}, \textit{Alternatively}, \textit{Maybe}, \textit{However}, which also occurs often in instruction with questions that are hard to solve.}
    \label{fig:rethinking_tokens}
\end{figure}

\begin{figure*}
    \centering
    \includegraphics[width=\linewidth]{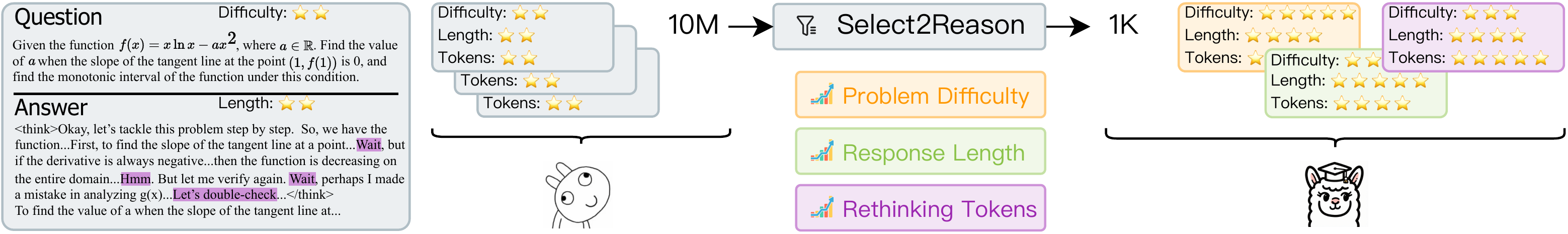}
    \caption{The brief pipeline of \textbf{\textsc{Select2Reason}} framework. With a large-scale instruction pool given, we select those data that can maximize the learning value of subsets via controlling the problem difficulty and reasoning trace length, which are motivated by the frequency of rethinking tokens during reasoning. The long-CoT reasoning ability of downstream model are activated after performing low-cost supervised fine-tuning on the instruction subset.}
    \label{fig:main}
\end{figure*}

A practical approach to activate long-CoT reasoning ability in pre-trained LLMs is to perform supervised fine-tuning (SFT) on instructions synthesized by strong LRMs. Open-source projects~\citep{openr1, OpenThoughts, Chinese-Data-Distill-From-R1} release over 100K such instructions respectively, yet large-scale SFT still entails significant costs. Recent work argues that the \textit{quality} of CoT data, rather than \textit{quantity} is more critical. LIMO~\citep{limo} applies multiple rounds of sampling and filtering over tens of millions of problems and employs expert-designed solutions to curate a compact yet high-quality dataset of 817 samples. S1~\citep{s1} depends heavily on API models and intricate data engineering pipelines tailored to optimize for quality, difficulty, and diversity, yielding 1k examples. Unfortunately, their metrics are based on qualitative heuristics without rigorous quantitative validation, and these carefully-curated pipelines are often not publicly available which impedes reproducibility and generalization.

Recently, research on instruction selection~\citep{alpagasus, DEITA, instag, autods, smalltolarge, IFD, selectit} has explored various aspects of data quality to automatically extract high-utility subsets from large instruction pools. However, the specific challenge of \textbf{\textit{instruction selection for long-CoT reasoning}} remains largely unaddressed. We investigate the features that may determine the quality of long-CoT instructions. The emergence of rethinking behaviors in long-CoT traces is regarded as an \textbf{\textit{aha moment}} for LRMs~\citep{deepseek}, indicating that the model learns to allocate more inference time toward self-correction and backtracking. Previous studies~\citep{logic-rl, demystifying} use the frequency of rethinking-related keywords as a proxy for reasoning quality, serving as a signal of the model’s internal thinking patterns. Similarly, s1~\citep{s1} implements budget forcing by appending extrapolation strings like \textit{Wait} to extend thinking process. We hypothesize that reasoning traces exhibiting more rethinking behaviors may serve as higher-quality instructions and offer greater training value. However, only qualitative keyword-based metrics cannot fully capture the complexity of reasoning patterns~\citep{simplerl}, highlighting the need for quantitative evaluation metrics.

We investigate common metrics that may lead to higher frequency of rethinking tokens in long-CoT reasoning trace, and statistical analysis presented in Figure~\ref{fig:rethinking_tokens} reveals that \textbf{\textit{longer}} reasoning traces exhibit more rethinking tokens in each step such as \textit{Wait}, \textit{Alternatively}, \textit{Maybe} and \textit{However}, which also occurs often in instruction with questions those are \textbf{\textit{hard}} to solve. It can be concluded that the \textbf{\textit{length of the reasoning trace}} in the response is a simple but tough-to-beat heuristic for selection. Furthermore, models trained on instruction subsets which are hard to solve by base model significantly outperform those trained on subsets with easy questions, aligning with the intuition in~\citep{limo, s1} that more challenging instructions provide greater learning value. However, the challenge of \textbf{\textit{automated, difficulty-aware}} instruction selection remains largely unaddressed.

To this end, we propose \textbf{\textsc{Select2Reason}}, a novel and efficient instruction-tuning data selection framework for Long-CoT reasoning. We leverage a difficulty-aware reward model to prioritize more challenging problems. Additionally, an instruction-response joint ranker that combines rankings based on difficulty and trace length controlled by a weighted scheme. We conduct extensive experiments across nine competition-level mathematical benchmarks to validate the efficacy. Built upon the OpenR1-Math dataset with 196K samples, \textbf{\textsc{Select2Reason}} selects the top 10\% instructions to fine-tune the Qwen2.5-Math-7B-Instruct model. Our method not only surpasses baselines but also matches or exceeds models trained on larger datasets, such as the OpenR1-Qwen-7B~\citep{openr1} with 94K samples and the DeepSeek-R1-Distill-Qwen-7B~\citep{deepseek} with 800K samples, demonstrating its effectiveness. Ablation studies highlight the scalability of our approach under varying data sizes. Additionally, model fine-tuned on high-quality data selected by \textsc{Select2Reason} conducts more efficient exploration using fewer thinking tokens when generating solution with stronger performance. \textsc{Select2Reason} demonstrates strong generalization by enabling low-cost transfer to other long-CoT reasoning instruction pools like Chinese-DeepSeek-R1-Distill dataset~\citep{Chinese-Data-Distill-From-R1} with 110K samples. Extensive case studies and visualizations support the effectiveness of our method.

Our contributions are summarized as follows: 1) We propose \textbf{\textsc{Select2Reason}}, a novel and efficient data selection framework for long-CoT instruction tuning. 2) We identify and validate key metrics including reasoning trace length and problem difficulty as strong heuristics for high-quality data selection. 3) We demonstrate remarkable performance on mathematical and broader reasoning benchmarks with lower training cost with verified scalability, robustness, and generalizability.

\section{Problem Definition}
\label{sec:problem}

\paragraph{Long-CoT Reasoning.} 
We focus on the capability of LRMs to generate long-CoT reasoning traces for solving questions with verifiable answers. Given a question $q \in \mathcal{Q}$ and a model $M$ parameterized by $\theta$, the model is expected to generate a reasoning trace $r$ including steps $\{s_1, s_2, \dots, s_n\}$, typically wrapped with \texttt{<think>} tokens, followed by a final answer $a \in \mathcal{A}$. Formally, the model outputs a pair $(r, a) \in \mathcal{R} \times \mathcal{A}$ such that:

{
\setlength{\abovedisplayskip}{6pt}
\setlength{\belowdisplayskip}{6pt}
\setlength{\abovedisplayshortskip}{4pt}
\setlength{\belowdisplayshortskip}{4pt}

\begin{equation}
\begin{aligned}
f_M(q) &= (r, a), \\
r &= \texttt{<think>} \; s_1, s_2, \dots, s_n \; \texttt{</think>}.
\end{aligned}
\end{equation}
}

The quality of CoT reasoning trace is often characterized by the emergence of human-like behaviors such as planning, verification, reflection, and backtracking. High-quality reasoning traces exhibit these traits to navigate complex problem spaces and are more likely to converge on correct solutions.

\paragraph{Instruction Selection.}
Instruction-tuning data selection aims to identify a optimal subset of reasoning instructions from a large instruction pool. Given a reasoning instruction dataset $\mathcal{D}_p = \{I_i\}_{i=1}^N$, where each instruction $I_i = (q_i, r_i \Vert a_i)$ includes a question, a reasoning trace, and a final answer, and a proposed evaluation metric suite $\boldsymbol{\pi} = \{\pi_1, \pi_2, \dots, \pi_k\}$ (e.g., quality, difficulty), our objective is to select a subset $\mathcal{D}_s \subseteq \mathcal{D}_p$ of size at most $K$ such that each selected instruction ranks among the top-$K$ under the metrics:

\begin{equation}
\mathcal{D}_s = \left\{I \in \mathit{Top}^{K}_{\boldsymbol{\pi}}(\mathcal{D}_p) \right\}.
\end{equation}

The fine-tuning objective is performed on $\mathcal{D}_s$ to update the model parameters $\theta$, thus minimizing the following negative log likelihood loss:

\begin{equation}
\min_{\theta} \; 
\mathcal{L}(\theta, \mathcal{D}_s)
= - \frac{1}{|\mathcal{D}_s|}
\sum_{(q, r, a) \in \mathcal{D}_s}
\log p_{\theta}([r; a] \mid q).
\end{equation}

\section{Exploration on Data Features}
\label{sec:rethinking}
In this section, we examine several metrics that may influence the frequency of rethinking tokens and conduct preliminary experiments to assess whether the metrics correlate with performance improvements, providing insights for selecting high-quality long-CoT instructions for fine-tuning.

\begin{figure}
    \centering
    \includegraphics[width=\linewidth]{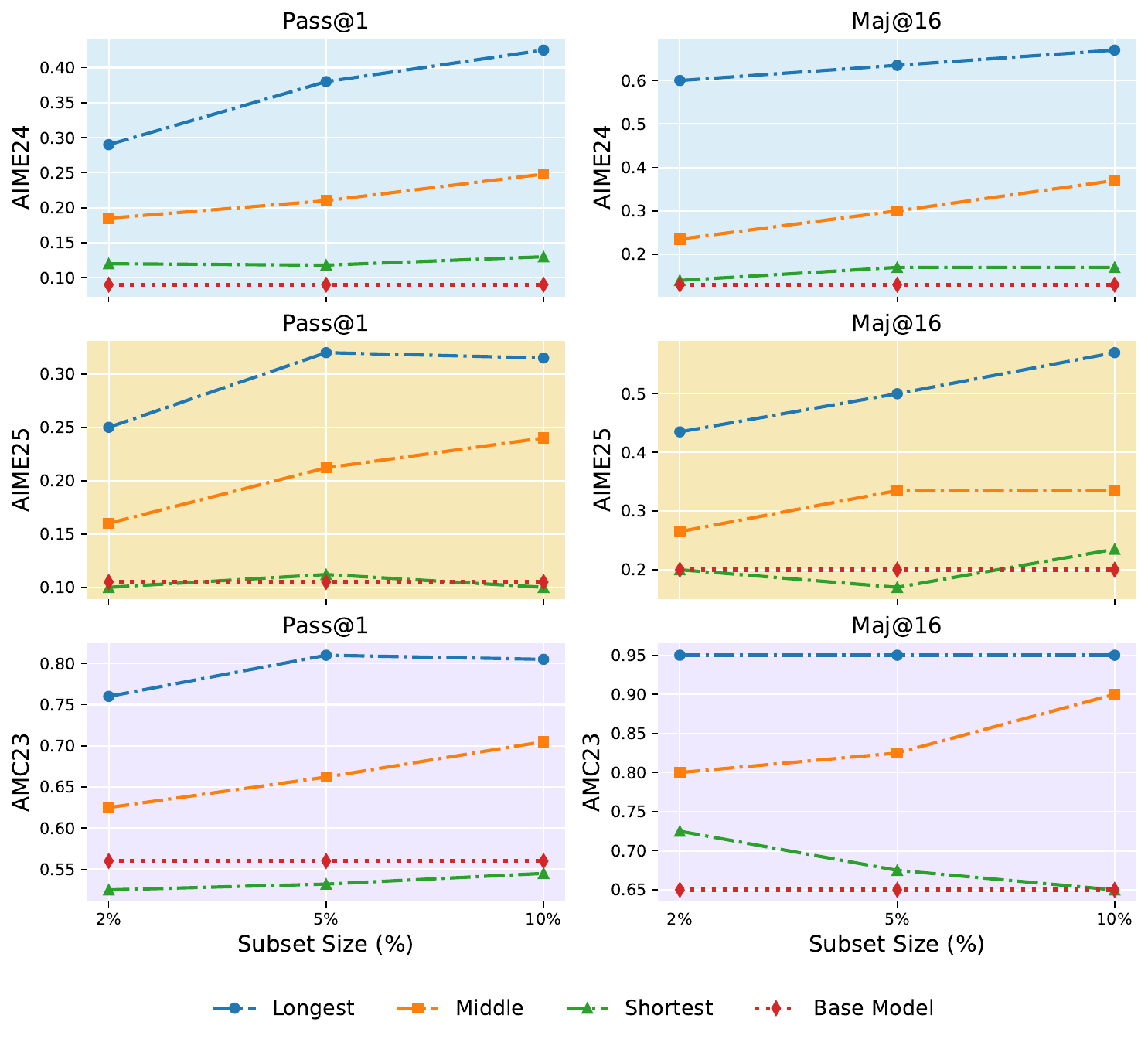}
    \caption{Performance across three expert-level mathematical benchmarks, using instruction subsets selected based on the length of reasoning traces, which are divided into the longest, the shortest and the middle.}
    \label{fig:trace_length}
    \vspace{-10pt}
\end{figure}

\begin{figure*}[t]
\centering
\scalebox{0.85}{
\begin{tcolorbox}[center,breakable,title=\Large\centering{Length of Reasoning Traces}]
\columnseprule=0.6pt
{\fontsize{9.0pt}{11.0pt}\selectfont
\begin{multicols}{2}

\begin{center}
{\large{\textbf{Long Reasoning Trace}}}
\end{center}
\textbf{Q:} Given the function \( f(x) = x\ln x - ax^2 \), where \( a \in \mathbb{R} \). Find the value of \( a \) when the slope of the tangent line at the point \( (1, f(1)) \) is 0, and find the monotonic interval of the function under this condition.\\
\textbf{A:}
\textit{<think>}\\
\textcolor{Green}{\textbf{[Problem Definition.]}}\\
Okay, let's tackle this problem step by step. So, we have the function \( f(x) = x \ln x - a x^2 \)...\\
\textcolor{Green}{\textbf{[Task Decomposition and Planning.]}}\\
...First, to find the slope of the tangent line at a point, we need the derivative of \( f(x) \)... Then, the derivative of \( -a x^2 \) is \( -2a x \)...\\
\textcolor{Green}{\textbf{[Self Reflection.]}}\\
...Wait, but if the derivative is always negative except at \( x=1 \) where it's zero, then the function is decreasing on the entire domain \( (0, \infty) \). Is that possible?...\\
\textcolor{Green}{\textbf{[Backtracking.]}}\\
...Hmm. But let me verify again. Wait, perhaps I made a mistake in analyzing \( g(x) \). Let's double-check...\\
\textit{</think>}\\
\textcolor{Green}{\textbf{[Final Answer.]}}\\
1. To find the value of \( a \) when the slope of the tangent line at...\\

\columnbreak

\begin{center}
{\large{\textbf{Short Reasoning Trace}}}
\end{center}
\textbf{Q:} Calculate the sum of the numbers 6 and 4. Add the number 30 to this sum.

\textbf{A:}
\textit{<think>}\\
\textcolor{Red}{\textbf{[No Self Reflection.]}}\\
...Let me do that step by step. 6 plus 4... that should be 10, right? Yeah, 6 plus 4 equals 10.\\
\textcolor{Red}{\textbf{[No Backtracking.]}}\\
...So putting it all together: first sum 6 and 4 to get 10, then add 30 to get 40. I don't think I made any mistakes here.\\
\textit{</think>}\\
\textcolor{Red}{\textbf{[Final Answer.]}} \\
The sum of 6 and 4 is calculated first: \( 6 + 4 = 10 \). Next, add 30 to ...\\

\begin{center}
\vspace{-20pt}
\rule{0.9\linewidth}{0.6pt}\\[0.6em]
{\large{\textbf{No Reasoning Trace}}}
\end{center}
\textbf{Q:} \( x, y \in \mathbb{R}^{+}, 3x + 2y = 12 \), maximize \( xy \)\\
\textbf{A:}
\textit{<think>}
\textbackslash n\textbackslash n\\
\textcolor{Red}{\textbf{[Thinking Terminate.]}}\\
\textit{</think>}\\
\textcolor{Red}{\textbf{[Final Answer.]}}\\
Given the constraint \( 3x + 2y = 12 \), we aim to maximize...\\
\vspace{1em}

\end{multicols}
}
\end{tcolorbox}
}
\caption{Comparison of \textbf{Instructions with Varying Reasoning Trace Lengths}. Long reasoning trajectories incorporate more human-like behaviors such as reflection, backtracking, and planning, which serve as higher-quality supervision signals during fine-tuning. In contrast, short traces often omit substantive decision-making steps and explicitly bypass reasoning by using empty constructs like \texttt{<think>}\textbackslash n\texttt{</think>}, rendering them ineffective.}
\label{fig:instruction-case}
\end{figure*}

\paragraph{Reasoning Traces with Varying Length.} We present the first systematic evaluation of how the length of the reasoning trace impacts instruction selection efficacy. Specifically, we sort the full set $\mathcal{D}_p$ by the reasoning trace length $r$, and construct subsets $\mathcal{D}_L^k$, $\mathcal{D}_M^k$, and $\mathcal{D}_S^k$ corresponding to the top-$k\%$ longest, middle, and shortest traces, respectively, for $k \in \{2, 5, 10\}$. As shown in Figure~\ref{fig:trace_length}, models fine-tuned on $\mathcal{D}_L^k$ consistently outperform those trained on $\mathcal{D}_M^k$ and $\mathcal{D}_S^k$ across different dataset sizes. Notably, $\mathcal{D}_M^k$ also yields positive gains over $\mathcal{D}_S^k$, highlighting a strong correlation between trace length and model improvement. While both $\mathcal{D}_L^k$ and $\mathcal{D}_M^k$ demonstrate scalable benefits with increasing subset size, the performance of model trained on $\mathcal{D}_S^k$ remains marginal or even negative. In Figure~\ref{fig:instruction-case}, long reasoning traces incorporate more rethinking behaviors such as reflection, backtracking, and planning, and serve as higher-quality supervision signals. In contrast, short traces often omit substantive decision-making steps and, in some cases, explicitly bypass reasoning by using empty constructs like \texttt{<think>}\textbackslash n\texttt{</think>}, rendering them ineffective. Figure~\ref{fig:rethinking_tokens} further confirm this point that longer reasoning traces exhibit a higher frequency of reflective steps that begin with patterns such as \textit{Wait}, \textit{Alternatively} or \textit{Maybe}. Motivated by these findings, we adopt \textbf{the longest reasoning traces as a simple, effective, and low-cost heuristic for data selection}, thereby avoiding the overhead of the reliance on costly human expert annotations~\citep{limo, s1}.

\begin{figure}
  \centering
  \includegraphics[width=\linewidth]{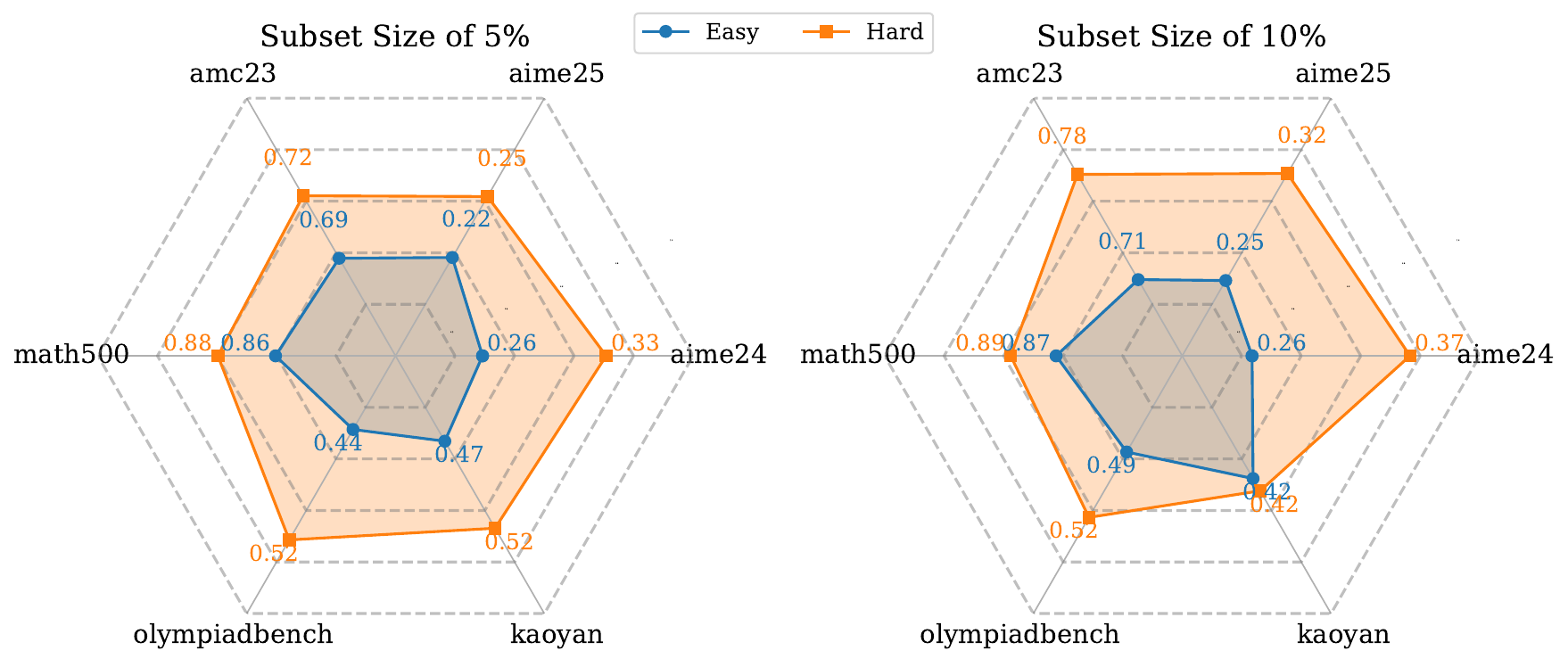}
  \caption{Pass@1 across six math benchmarks. Easy and hard examples for base model are selected for training, and the latter brings more learning value.}
  \label{fig:diff_radar}
  \vspace{-10pt}
\end{figure}

\paragraph{Difficulty of Question.} Difficulty as a criterion for data selection is applied with the prevailing intuition being that more challenging questions offer higher learning value. The model performs inference over a sampled subset of instructions, and label instances as easy or hard based on whether the model successfully solves the question. This yields $\mathcal{D}_E^k$ (easy) and $\mathcal{D}_H^k$ (hard), for $k \in \{5, 10\}$. As shown in Figure~\ref{fig:diff_radar}, models fine-tuned on $\mathcal{D}_H^k$ consistently outperform those trained on $\mathcal{D}_E^k$ across six mathematical benchmarks in two data scales. This aligns with prior findings and our hypothesis that instructions with challenging question facilitate stronger supervision with richer training signals. Existing work~\citep{limo, s1} has not established quantitative criteria for identifying the difficulty of questions, thereby limiting the prioritization of hard problems and \textbf{leaving the challenge of automated, difficulty-aware} instruction selection unaddressed.

\paragraph{Diversity.} Diversity has long been a central theme in instruction selection~\citep{DEITA, instag, SAE, CaR}. However, simple heuristics such as uniformly sampling from each domain offers no clear advantage over random selection for long-CoT reasoning instructions~\citep{s1}. To further examine the role of diversity, we leverage \textbf{\textit{metadata}} from the Open-R1-Math instruction set where problems are categorized into topics like \textit{algebra, geometry, calculus} and so on. We sample a domain-balanced subset $\mathcal{D}_D^k$ and compare it against a randomly sampled one $\mathcal{D}_R^k$ of the same size. As shown in Figure~\ref{fig:diversity}, the model fine-tuned on $\mathcal{D}_D^k$ does not exhibit significant performance gains over the baseline, and in some cases, such as Major@16 on AMC 23, the performance curves nearly overlap. These results suggest that diversity may not contribute meaningfully for instruction selection in long-CoT reasoning, serving as a baseline only.

\section{\textsc{Select2Reason}}
\label{sec:method}

We propose \textsc{Select2Reason}, an efficient instruction-tuning data selection method for long-CoT reasoning. Our method selects a compact yet high-utility subset from a large instruction pool by jointly modeling (i) question difficulty quantified by a reward model-as-a-judge and (ii) the effective length of reasoning traces after normalization.





\subsection{Difficulty-aware Reward Model-as-Judge}

\paragraph{Rollout-based Difficulty Signal.}
Instead of relying on static or heuristic difficulty annotations,
we estimate question difficulty from the perspective of the base model.
For each question $q_i$, we perform $R$ Monte Carlo rollouts using the base model $M$, obtaining a set of candidate solutions
$\{(r_i^{(j)}, a_i^{(j)})\}_{j=1}^{R}$, where we define the empirical solving accuracy with ground truth given as:
\begin{equation}
\hat{p}_{\text{solve}}(q_i) = \frac{1}{R} \sum_{j=1}^{R}
\mathbb{I}[a_i^{(j)} = gt_i],
\end{equation}
and convert it into a difficulty score 
$\mathrm{diff}(q_i) = 1 - \hat{p}_{\text{solve}}(q_i)$,
where higher values indicate questions that are more challenging for the current model.

\paragraph{Reward Model Training and Inference.}
Given that each question $q_i$ in the supervision set already has an estimated
difficulty label $\mathrm{diff}(q_i)$ obtained from rollout-based
evaluation, we directly train a reward model to predict this continuous
difficulty value. We construct the supervision set
$\mathcal{D}_s = \{(q_i, \mathrm{diff}(q_i))\}$,
where the reward model takes the question as input and learns to regress
toward the target difficulty score. Formally, the model parameters $\theta$
are optimized by minimizing the negative log-likelihood:

\begin{equation}
\min_{\theta} \;
\mathcal{L}(\theta, \mathcal{D}_s)
=
-
\frac{1}{|\mathcal{D}_s|}
\sum_{q_i \in \mathcal{D}_s}
\log p_{\theta}\!\left(\mathrm{diff}(q_i) \mid q_i\right),
\end{equation}

where $p_{\theta}(\mathrm{diff}(q_i) \mid q_i)$ denotes the predicted
probability density assigned by the reward
model to the target difficulty value. 

After training, the reward model serves as an efficient
\textit{reward-model-as-judge} for difficulty estimation. During inference, each instruction in the pool $\mathcal{D}_p$ is scored by the trained model to obtain a calibrated, continuous difficulty score.

\subsection{Normalization of Reasoning Trace}

Raw length of long reasoning traces may be inflated by repeated reasoning steps,
e.g., verbatim restatements or looping thoughts. Given a reasoning trace
$r_i = \{s_1, s_2, \dots, s_{n_i}\}$,
where each $s_k$ corresponds to a discrete reasoning step,
we construct a normalized trace
by removing steps that are exact duplicates of any previous step.
Formally, a step $s_k$ is discarded if $\exists \; j < k \;\; \text{s.t.} \;\; s_k = s_j$. Let $\tilde{r}_i$ denote the resulting deduplicated reasoning trace,
and the normalized reasoning length is defined as
$\mathrm{len}_{\text{norm}}(r_i) = |\tilde{r}_i|$. This normalization penalizes trivial repetition
while preserving genuinely distinct reasoning steps.

\subsection{Joint Ranker for Instruction Utility}
While we now have an efficient method to score questions via difficulty and responses via reasoning trace length, combining them in a principled manner remains a challenge. Inspired by prior work on multi-criteria ranking~\citep{instructionmining, QDIT}, we aggregate rankings using a weighted scheme. Let $\mathrm{rank}_d(I_i)$ and $\mathrm{rank}_l(I_i)$ denote the rankings of instruction $I_i$ by question difficulty and reasoning trace length, we define the joint ranking with a weighted scheme as:

\begin{equation}
\mathit{rank}_j(I_i) = w \cdot \mathit{rank}_d(I_i) + (1 - w) \cdot \mathit{rank}_l(I_i),
\end{equation}

where a weighting factor $w \in [0, 1]$ controls the trade-off between rankings by difficulty and trace length. The final selected subset by out methods for training downstream base model is:

\begin{equation}
\mathcal{D}_{\textsc{Select2Reason}} = \left\{I \in \mathit{Top}^{K}_{\mathit{rank}_j}(\mathcal{D}_p) \right\}.
\end{equation}

\section{Experiment Results and Analysis}
\label{sec:analysis}

\begin{table*}
  \caption{Performance of \textsc{Select2Reason} on the \textit{OpenR1-Math-220k} pool across nine benchmarks using Pass@1 (P@1) and Majority@16 (M@16) as metrics. We incorporate two open-source models and ablation baselines for data selection like filtering randomly, by diversity, by reasoning trace length and by problem difficulty for reference.}
  \label{tab:main_results}
  \centering
  \tabcolsep 1pt
  \small
  \begin{tabular}{l | c | c | c | c | c | c | c | c | c |c|c|c|c}
    \toprule
    \textbf{Target Model} & \textbf{Data} & \multicolumn{2}{c|}{\textbf{AIME 24}} & \multicolumn{2}{c|}{\textbf{AIME 25}} & \multicolumn{2}{c|}{\textbf{AMC 23}} & \textbf{MATH} & \textbf{Olympiad} & \textbf{Kaoyan} & \textbf{GK 23} & \textbf{GK-Math} & \textbf{GK 24} \\
    \cmidrule(r){3-4} \cmidrule(r){5-6} \cmidrule(r){7-8} \cmidrule(r){9-9} \cmidrule(r){10-10} \cmidrule(r){11-11} \cmidrule(r){12-12} \cmidrule(r){13-13} \cmidrule(r){14-14}
    \textsc{Qwen2.5-Math-7B} & \textbf{Size} & P@1 & M@16 & P@1 & M@16 & P@1 & M@16 & P@1 & P@1 & P@1 & P@1 & P@1 & P@1 \\
    \midrule
    \textsc{Base Model} & - & 0.090 & 0.133 & 0.106 & 0.200 & 0.558 & 0.650 & 0.842 & 0.394 & 0.472 & 0.649 & 0.781 & 0.637 \\
    \midrule
    \textsc{R1-Distill-Qwen} & 800k & 0.544 & 0.833 & 0.417 & 0.600 & 0.895 & 0.950 & 0.896 & 0.551 & 0.618 & 0.810 & 0.880 & 0.692 \\
    \midrule
    \textsc{OpenR1-Qwen} & 94k & 0.460 & 0.700 & 0.317 & 0.467 & 0.823 & 0.950 & 0.906 & 0.526 & 0.492 & 0.795 & 0.843 & 0.714 \\
    \midrule
    \textsc{Full-Pool} & 196k & 0.465 & 0.700 & 0.352 & 0.600 & 0.816 & 0.950 & 0.894 & 0.560 & 0.382 & 0.800 & 0.783 & 0.615 \\
    \midrule
    \textsc{Random} & \multirow{5}{*}{10\%} & 0.331 & 0.600 & 0.267 & 0.367 & 0.753 & \textbf{0.950} & 0.878 & 0.510 & 0.467 & 0.740 & 0.789 & 0.626 \\
    \textsc{Diverse} & & 0.327 & \textbf{0.667} & 0.267 & 0.433 & 0.750 & \textbf{0.950} & 0.846 & 0.493 & 0.467 & 0.745 & 0.809 & 0.659 \\
    \textsc{Longest} & & 0.425 & \textbf{0.667} & 0.312 & \textbf{0.567} & 0.805 & \textbf{0.950} & 0.898 & 0.535 & 0.548 & 0.795 & \textbf{0.892} & \textbf{0.747} \\
    \textsc{Difficult} & & 0.410 & 0.633 & 0.312 & 0.433 & 0.787 & 0.925 & 0.886 & 0.530 & 0.533 & 0.787 & 0.866 & 0.703 \\
    \rowcolor{gray!20}\textsc{Select2Reason} & & \textbf{0.433} & \textbf{0.667} & \textbf{0.335} & \textbf{0.567} & \textbf{0.808} & \textbf{0.950} & \textbf{0.914} & \textbf{0.548} & \textbf{0.573} & \textbf{0.800} & \textbf{0.892} & 0.736 \\
    \bottomrule
  \end{tabular}
\end{table*}

\paragraph{Datasets and Experiment Settings.}
\label{sec:experiment}
We adopt \textit{OpenR1-Math-220k}~\citep{openr1} as the data pool, which is a large-scale instruction set for long-CoT reasoning distilled from DeepSeek-R1~\citep{deepseek}. We retained the portion containing the correct answers, approximately 196k samples. We employ \textit{Qwen2.5-Math-7B-Instruct}~\citep{qwenmath} as the backbone model. We adopt nine mathematics benchmark spanning multiple dimensions for evaluation, including three competition-level benchmarks which are AIME in 2024 \& 2025, and AMC in 2023, and six comprehensive benchmarks such as MATH-500~\citep{math500} and OlympiadBench~\citep{olympiadbench} for math reasoning, and GAOKAO in 2023 \& 2024~\citep{qwenmath}, GAOKAO MATH~\cite{qwenmath} and KAOYAN~\citep{limo} in Chinese math. More details are provided in Appendix~\ref{app:experiment} and ~\ref{app:evaluation}.

\begin{figure}
  \centering
  \includegraphics[width=\linewidth]{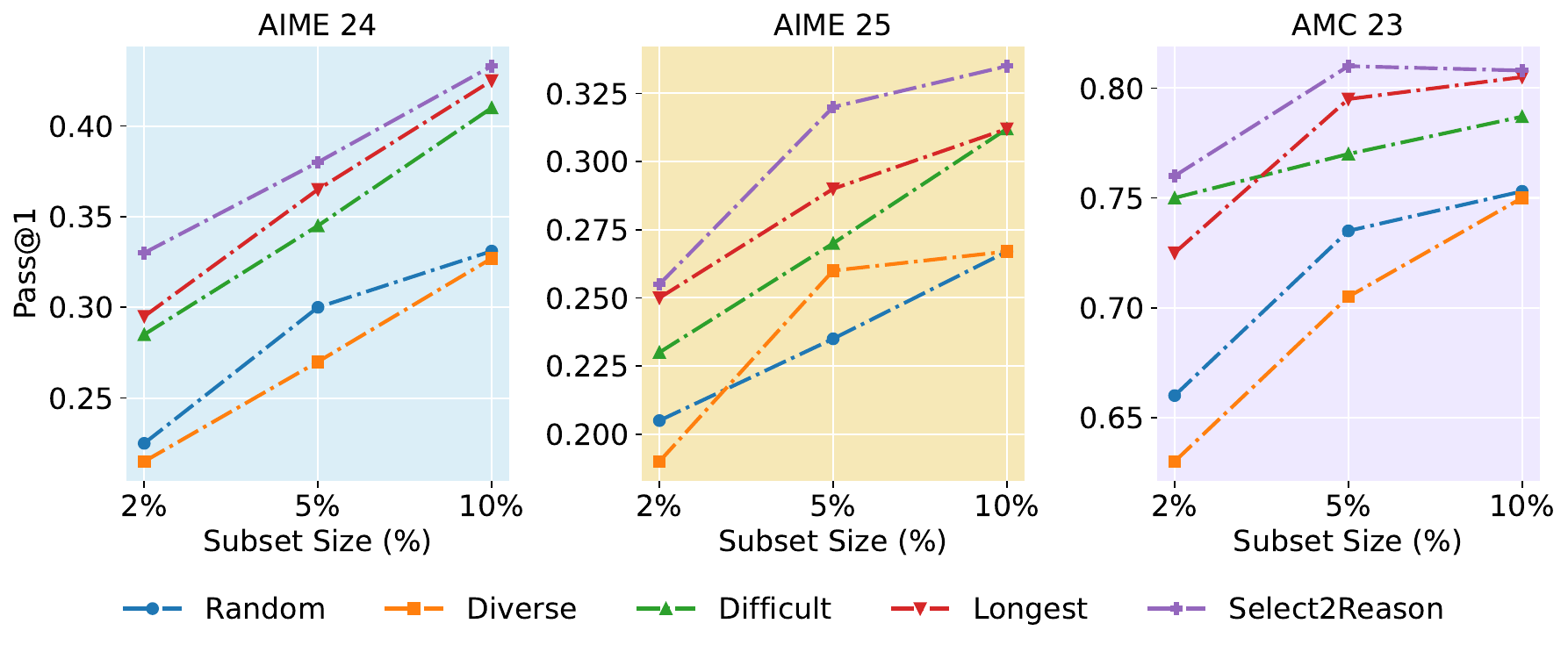}
  \caption{Performance across three benchmarks of baselines and our method in varying subset size.}
  \label{fig:scaling}
  \vspace{-10pt}
\end{figure}

\paragraph{Main Results.}
Table~\ref{tab:main_results} presents the performance of \textsc{Select2Reason} across nine mathematics benchmarks. We compare against DeepSeek-R1-Distill-Qwen-7B~\citep{deepseek}, which is trained on 800k instructions, and OpenR1-Qwen-7B~\citep{openr1} trained on 94k instructions. We evaluate four baseline selection strategies: \textit{Random}, which samples instructions uniformly from the pool; \textit{Diverse}, which performs clustering and balanced sampling based on category metadata; \textit{Longest}, which selects instructions with the longest reasoning traces; and \textit{Difficult}, which chooses top-ranked samples according to our difficulty quantifier. \textsc{Select2Reason} outperforms all baselines on most datasets, consistently achieving higher Pass@1 scores and matching the strongest baselines on Maj@16. Furthermore, the model trained on the subset filtered by \textsc{Select2Reason} surpasses both \textit{Full-pool} models and open-source models on MATH-500. Notably, our method maintains a strong lead over \textit{Full-pool} training on nearly all comprehensive math benchmarks, with only a slight performance drop against the \textit{Full-pool} on OlympiadBench. This reflect a limitation in the generalization of \textit{Full-pool} setting, whereas training with a smaller, high-quality subset yields superior performance.

\begin{figure}
  \centering
  \includegraphics[width=\linewidth]{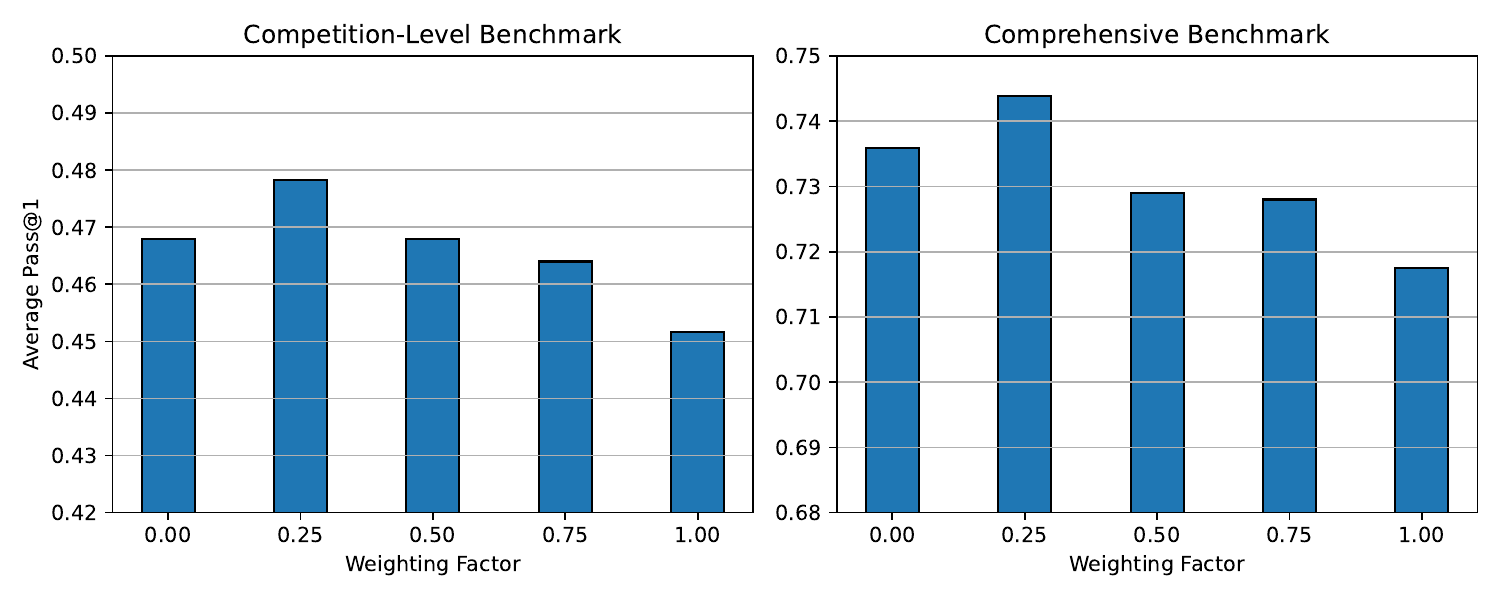}
  \caption{Average Pass@1 by adjusting the weighting factor of joint ranker in \textsc{Select2Reason}.}
  \label{fig:weight_scheme}
  \vspace{-12pt}
\end{figure}

\paragraph{Performance of \textsc{Select2Reason} under different hyperparameter settings.}

\begin{table*}
  \caption{Generalizability of \textsc{Select2Reason} on the \textit{Chinese-DeepSeek-R1-Distill-data} pool with 110k samples.}
  \label{tab:chinese_results}
  \centering
  \tabcolsep 1pt
  \small
  \begin{tabular}{l | c | c | c | c | c | c | c | c | c | c | c | c | c}
    \toprule
    \textbf{Target Model} & \textbf{Data} & \multicolumn{2}{c|}{\textbf{AIME 24}} & \multicolumn{2}{c|}{\textbf{AIME 25}} & \multicolumn{2}{c|}{\textbf{AMC 23}} & \textbf{MATH} & \textbf{Olympiad} & \textbf{Kaoyan} & \textbf{GK 23} & \textbf{GK-Math} & \textbf{GK 24} \\
    \cmidrule(r){3-4} \cmidrule(r){5-6} \cmidrule(r){7-8} \cmidrule(r){9-9} \cmidrule(r){10-10} \cmidrule(r){11-11} \cmidrule(r){12-12} \cmidrule(r){13-13} \cmidrule(r){14-14}
    \textsc{Qwen2.5-Math-7B} & Size & P@1 & M@16 & P@1 & M@16 & P@1 & M@16 & P@1 & P@1 & P@1 & P@1 & P@1 & P@1 \\
    \midrule
    \textsc{Base Model} & - & 0.090 & 0.133 & 0.106 & 0.200 & 0.558 & 0.650 & 0.842 & 0.394 & 0.472 & 0.649 & 0.781 & 0.637 \\
    \midrule
    \textsc{Full-Pool} & 110K & 0.181 & 0.267 & 0.158 & 0.300 & 0.633 & 0.800 & 0.798 & 0.367 & 0.412 & 0.668 & 0.718 & 0.626 \\
    \midrule
    \textsc{Random} & \multirow{5}{*}{10\%} & 0.181 & 0.200 & 0.140 & 0.167 & 0.620 & 0.725 & 0.840 & 0.431 & 0.372 & \textbf{0.701} & 0.775 & 0.626 \\
    \textsc{Diverse} &  & 0.176 & 0.233 & 0.144 & 0.233 & 0.618 & 0.750 & 0.822 & 0.416 & 0.377 & 0.692 & 0.770 & 0.641 \\
    \textsc{Longest} &  & 0.221 & 0.367 & 0.173 & 0.233 & 0.656 & \textbf{0.850} & 0.846 & \textbf{0.459} & 0.457 & 0.688 & 0.821 & \textbf{0.703} \\
    \textsc{Difficult} &  & \textbf{0.258} & \textbf{0.400} & 0.194 & 0.267 & 0.627 & 0.800 & 0.848 & 0.412 & \textbf{0.462} & 0.691 & 0.795 & 0.681 \\
    \rowcolor{gray!20}\textsc{Select2Reason} & & 0.242 & \textbf{0.400} & \textbf{0.206} & \textbf{0.367} & \textbf{0.689} & 0.825 & \textbf{0.860} & 0.450 & \textbf{0.462} & 0.699 & \textbf{0.840} & \textbf{0.703} \\
    \bottomrule
  \end{tabular}
\end{table*}

Figure~\ref{fig:scaling} presents a statistical analysis of performance variation for both the baselines and \textsc{Select2Reason} across different subset sizes (2\%, 5\%, and 10\%) on three datasets. \textsc{Select2Reason} consistently maintains a leading advantage as the subset size increases, where the performance generally improves in a stable manner. Another critical hyperparameter is the weighted factor $w \in [0,1]$ used in the joint ranker. Figure~\ref{fig:weight_scheme} shows the sensitivity analysis. When $w=0$, the joint ranker degenerates to the length-based ranker; when $w=1$, it becomes equivalent to the difficulty-based ranker. The best performance is achieved at $w=0.25$, where the model fine-tuned on the top 10\% subset reaches highest accuracy. This indicates that the joint ranker achieves an effective balance in controlling the trade-off.

\begin{figure}
  \vspace{-15pt}
  \includegraphics[width=\linewidth]{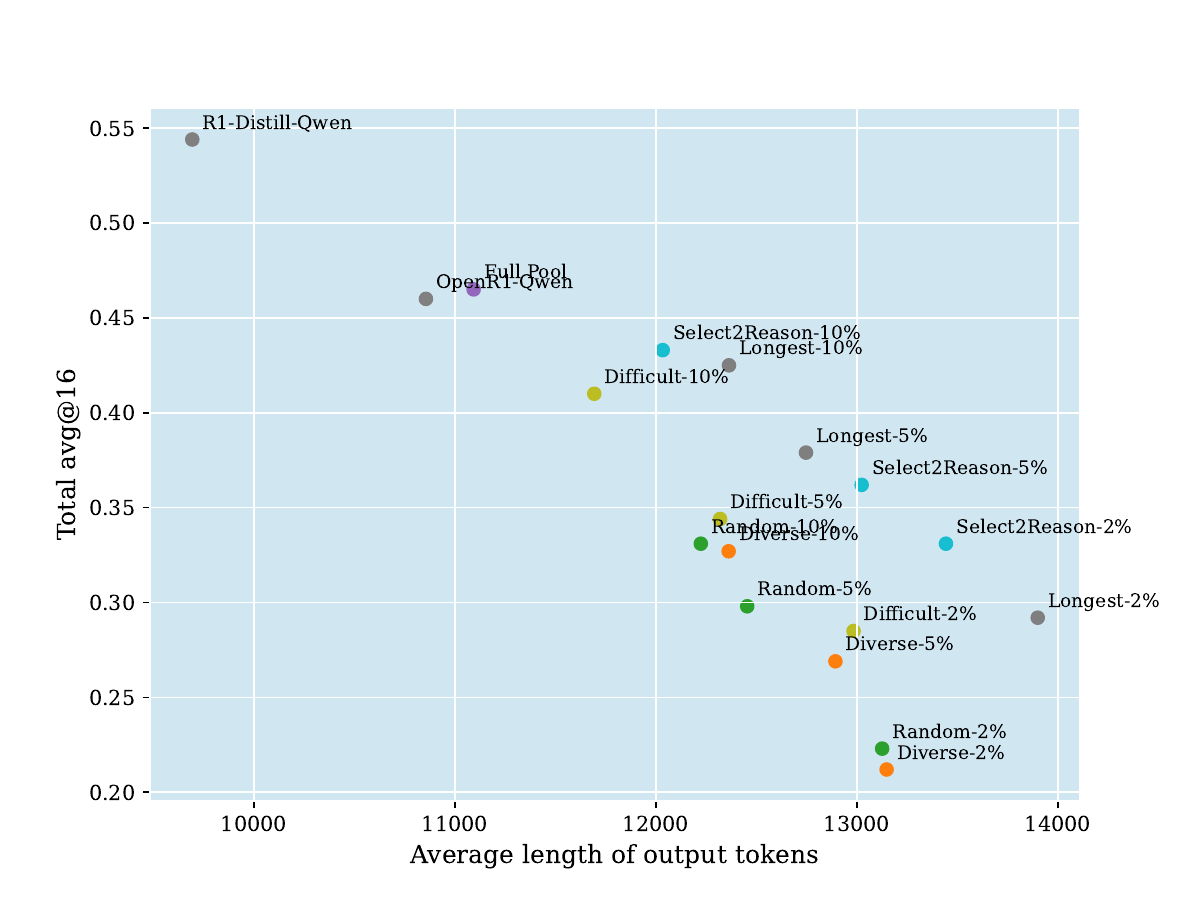}
  \caption{The relationship between models' performance and output length on AIME 25. Model trained with high-quality data filtered by \textsc{Select2Reason} demonstrate a better reasoning efficiency.}
  \label{fig:performance_tokens}
  \vspace{-10pt}
\end{figure}

\paragraph{\textbf{\textsc{Select2Reason} improves long-CoT reasoning efficiency by sampling high-quality data.}} 
The relationship between performance and output tokens on AIME 25 is illustrated in Figure~\ref{fig:performance_tokens}. In contrast to the increasing response lengths observed during conducting pure RL~\citep{deepseek}, SFT exhibits a different distribution, where models with stronger performance tend to generate shorter outputs. This suggests that \textit{models produce more efficient exploratory solutions when long-CoT reasoning is effectively activated}. Further analysis is presented in Figure~\ref{fig:tokens_on_aime}, which shows the frequency of rethinking tokens used by fine-tuned models on AIME 25. The model trained on \textsc{Select2Reason} subsets consistently uses fewer rethinking tokens across all data sizes, supporting our hypothesis that it enables more efficient reasoning. A case study in Figure~\ref{fig:case_efficient} illustrated that a model trained on low-quality instructions attempts to use a large number of rethinking tokens during inference, but exhibits limited effective reflection. 
In comparison, a model trained on higher-quality instructions corrects its reasoning path and reaches the correct solution with fewer rethinking steps.

\begin{table}
  \caption{Comparison of Select2Reason (S2R) with prior works like S1.1~\citep{s1} and LIMO~\citep{limo} under different settings.}
  \label{tab:compare_prior_work}
  \centering
  \tabcolsep 4.5pt
  \scriptsize
  \begin{tabular}{l | c | c | c | c}
    \toprule
    \textbf{Model} & \textbf{Data Size} & \textbf{AIME 24} & \textbf{AIME 25} & \textbf{AMC 23} \\
    \midrule
    \textsc{Base Model} & - & 0.090 & 0.106 & 0.558 \\
    \midrule
    \multicolumn{5}{c}{\textit{Individual Corpus}} \\
    \midrule
    Qwen2.5-\textsc{S2R} & 982 & \textbf{0.283} & \textbf{0.237} & \textbf{0.728} \\
    Qwen2.5-\textsc{S1.1} & 1k & 0.225 & 0.198 & 0.669 \\
    Qwen2.5-\textsc{LIMO} & 871 & 0.206 & 0.210 & 0.627 \\
    \midrule
    \multicolumn{5}{c}{\textit{Full Corpus of S1}} \\
    \midrule
    Qwen2.5-\textsc{S1.0 (Full)} & 59k & 0.224 & 0.169 & 0.588 \\
    Qwen2.5-\textsc{S1.0 (1k)} & 1k & 0.202 & 0.154 & \textbf{0.614} \\
    Qwen2.5-\textsc{S1.0-S2R} & 1k & \textbf{0.238} & \textbf{0.177} & 0.606 \\
    \bottomrule
  \end{tabular}
\end{table}

\paragraph{\textsc{Select2Reason} yields higher quality comparing with datasets in prior methods.}
We conduct a comparative analysis with \textsc{LIMO}~\citep{limo} and \textsc{S1}~\citep{s1} in data quality.
Specifically, we select approximately 1k long-CoT instructions from each synthesized data pool to evaluate performance across five benchmarks. 
As shown in Table~\ref{tab:compare_prior_work}, the model trained on instructions selected by \textsc{Select2Reason} consistently outperforms those trained on data selected by \textsc{LIMO} and \textsc{S1.1}. Furthermore, we apply \textsc{Select2Reason} to the full instruction pool used by \textsc{S1.0}, which comprises 59k examples with Gemini Flash Thinking responses (the full pool used by LIMO is not publicly available) to select a 1k subset, and again observe improved performance over the original selected 1k subset of \textsc{S1.0}.

\begin{figure*}
  \centering
  \includegraphics[width=\linewidth]{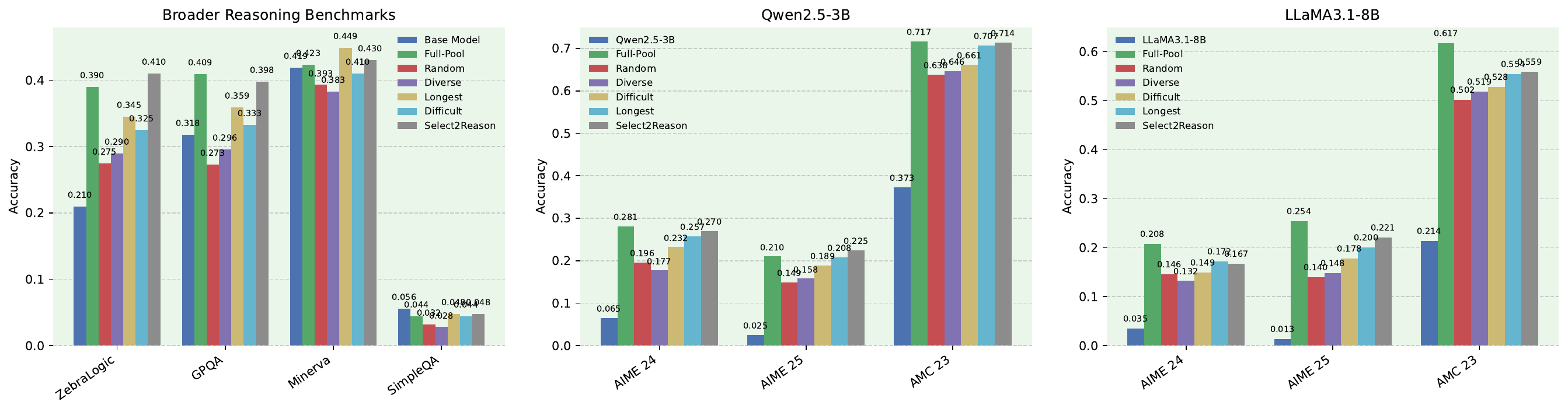}
  \caption{(a) Generalization of \textsc{Select2Reason} in broader reasoning tasks, including logical reasoning, scientific QA and commonsense QA. (b)(c) Generalization of \textsc{Select2Reason} across language models with smaller size like \textit{Qwen2.5-3B-Instruct} and various language models like \textit{LLaMA3.1-8B-Instruct}.}
  \label{fig:generalization}
\end{figure*}

\paragraph{\textsc{Select2Reason} demonstrates strong generalization capabilities by enabling low-cost transfer to other Long-CoT reasoning instruction pools.} 
To assess the generalizability of \textsc{Select2Reason} on \textit{Chinese-DeepSeek-R1-Distill-data}, we directly apply the joint ranker trained on OpenR1-Math-220k. Results in Table~\ref{tab:chinese_results} show that fine-tuning model on only the top 10\% subset selected by \textsc{Select2Reason} outperforms baselines. Notably, since this data pool contains a large proportion of generic, non-reasoning instructions, we conclude that this dilutes the model’s ability to acquire strong reasoning capabilities. 
Case studies of joint ranking are presented in Appendix~\ref{sec:case_chinese}, despite not being trained on this specific instruction pool, the joint ranker still successfully identifies high-quality reasoning instructions, demonstrating the notable generalizability of \textsc{Select2Reason}.

\begin{table}
  \caption{Comparison of computational overhead in data selection between trained with full-pool dataset and high-utility subset filtered by \textsc{Select2Reason}.}
  \label{tab:compute_cost}
  \centering
  \tabcolsep 4pt
  \scriptsize
  \begin{tabular}{l | c | c | c | c}
    \toprule
    \textbf{Type} & \textbf{Data Size} & \textbf{Strategy} & \textbf{Time Cost} & \textbf{GPU Usage} \\
    \midrule
    \multicolumn{5}{c}{\textit{Open-R1-Math-220k}} \\
    \midrule
    Full Pool & 196k & Full-SFT & 40 hours & 8×40GB \\
    Reward Model & 4k & LoRA-SFT & 10 mins & 8×40GB \\
    Difficulty Scoring & 196k & Inference & 11 mins & 2×40GB \\
    Length Ranking & 196k & - & - & - \\
    Select2Reason & 19.6k & Full-SFT & 10 hours & 8×40GB \\
    \midrule
    \multicolumn{5}{c}{\textit{Chinese-R1-Distill-110k}} \\
    \midrule
    Full Pool & 110k & Full-SFT & 10 hours & 8×40GB \\
    Reward Model & - & Already & - & - \\
    Difficulty Scoring & 110k & Inference & 3 mins & 2×40GB \\
    Length Ranking & 110k & - & - & - \\
    Select2Reason & 11k & Full-SFT & 3 hours & 8×40GB \\
    \bottomrule
  \end{tabular}
\end{table}

\paragraph{\textsc{Select2Reason} demonstrates robust generalization across domains and model scales.}
Beyond the mathematical domain, we extend our evaluation to broader reasoning tasks, including logical inference, scientific QA, and commonsense reasoning. As summarized in Figure~\ref{fig:generalization} (a), \textsc{Select2Reason} consistently achieves superior performance compared to baseline strategies, particularly on benchmarks such as ZebraLogic and GPQA, where long-CoT reasoning is essential. Furthermore, to assess robustness across model families and scales, we fine-tune two additional open-source models, Qwen2.5-3B-Instruct~\citep{qwen2.5} and LLaMA-3.1-8B-Instruct~\citep{llama}. The results in Figure~\ref{fig:generalization} (b)(c) confirm that \textsc{Select2Reason} maintains its advantage across both smaller-scale models and different architectures. Specifically, while absolute performance decreases with model size, the relative improvements over baselines remain consistent, validating that the effectiveness of our method is not confined to a single model family or scale.

\paragraph{\textsc{Select2Reason} achieves significant training efficiency with minimal selection overhead.} We conduct a detailed cost-benefit analysis of \textsc{Select2Reason} to assess its computational efficiency. As shown in Table~\ref{tab:compute_cost}, the total overhead introduced by the selection process is minimal compared to the cost of training on the full instruction pool. Notably, this results in a nearly 75\% reduction in training time without compromising performance. Additionally, when applied to a new data pool, the judge model generalizes effectively without retraining, and the inference stage completes within 3 minutes. This demonstrates the transferability and amortized cost of the pipeline.

\section{Conclusion}
\label{sec:conclusion}
While recent large reasoning models exhibit remarkable long-CoT reasoning abilities, effective instruction selection remains an underexplored challenge. Our study identifies reasoning trace length and problem difficulty as strong and quantifiable heuristics for high-quality data selection. Building on these insights, we introduce \textsc{Select2Reason}, an novel and efficient instruction-tuning data selection framework. Extensive experiments demonstrates that models trained on selected subsets achieve superior performance with significantly less data, paving the way for cost-effective and high-quality long-CoT instruction tuning.

\section*{Limitations}
Despite the promising results of \textsc{Select2Reason} in enhancing long-CoT reasoning capabilities, several limitations remain. First, due to constraints in computational resources and training costs, our experiments are primarily conducted on medium-scale models, and the scalability of our method to larger models remains to be explored. Second, the current study relies on existing instruction datasets, while automated instruction evolution strategies to improve data quality are yet to be developed. Finally, although our analysis reveals correlations between reasoning trace length, problem difficulty, and rethinking behaviors, the interpretability of how long-CoT capabilities are activated and how reflective reasoning emerges during SFT remains an open question for future work.
\bibliography{custom}

\clearpage
\appendix
\onecolumn
\section*{Appendix}

\section{Related Work}
\label{sec:related}

\paragraph{Reasoning of Large Language Models.} LLMs demonstrate notable chain-of-thought (CoT)~\citep{cot} reasoning capabilities that are critical for addressing complex tasks such as mathematical problem solving, coding, and logical inference~\citep{reasonsurvey, cotsurvey, system2survey}. Various efforts aim to enhance LLMs' reasoning through different training stages. Previous works~\citep{codellama, deepseekmath} reinforce models to memorize reasoning patterns by injecting high-quality knowledge and rationales during pre-training. Furthermore, carefully curated datasets~\citep{metamath, cotcollection, logicot} significantly boost complex reasoning performance through fine-tuning~\citep{RFT}. Some studies focus on scaling inference-time computation~\citep{scalingtest}, such as employing self-consistency or reward-based verifiers to validate outcome or process on sampled candidate solutions~\citep{selfconsistency, stepbystep, mathshepherd}. Recently, researchers have observed planning and self-reflection behaviors in long-CoT responses of large reasoning models such as OpenAI-o1~\citep{o1}, DeepSeek-R1~\citep{deepseek}, Kimi-1.5~\citep{kimi}, QwQ~\citep{qwq} and Gemini Thinking~\citep{gemini}, symbolizing a major breakthrough in complex reasoning. Open community projects~\citep{openr1, OpenThoughts} contribute by organizing synthetic datasets and distilling reasoning abilities from DeepSeek-R1 into smaller LLMs. 

\paragraph{Instruction-Tuning Data Selection.}
Instruction-tuning data selection aims to identify high-utility subsets from large instruction pools to improve model performance and alignment. Early efforts emphasized human expert curation~\citep{lima}, while recent work has explored automated selection using various metrics. GPT-based judgments of instruction-response quality are commonly used~\citep{alpagasus, QDIT, selectit, autods, crowdselect}, often enhanced with diversity signals~\citep{DEITA, instag, iterselecttune, SAE, mig}. Several studies leverage model-internal features such as loss~\citep{IFD, mods, nuggets, recost}, gradients~\citep{less, g-dig, tagcos}, perplexity~\citep{superfiltering, LPapp}, and linguistic features~\citep{instructionmining, long} to assess sample utility. Techniques like weak-to-strong supervision~\citep{smalltolarge, superfiltering, LPapp} and expert preference-aligned scoring~\citep{CaR} further enrich the selection space. With the advent of large reasoning models, LIMO~\citep{limo}, LIMR~\citep{limr} and s1~\citep{s1} observe that training with a few carefully crafted reasoning examples can achieve remarkable performance, highlighting the necessity of efficient minimal supervision. However, automatic selection of long-CoT reasoning instructions remains unexplored, where criteria have not been designed and verified.

\newpage
\section{Experimental Settings}
\label{app:experiment}
We adopt \textbf{OpenR1-Math-220k}~\citep{openr1} as the data pool, which is a large-scale instruction set for long-CoT reasoning distilled from DeepSeek-R1~\citep{deepseek} using math problems from NuminaMath~\citep{numinamath}. We retain part of them that lead to a correct answer about \textbf{196k}. We also adopt \textbf{Chinese-DeepSeek-R1-Distill-data}~\citep{Chinese-Data-Distill-From-R1}, a open-source dataset containing 110k Chinese instructions spanning mathematics, STEM, and general domains, with Long-CoT responses generated by DeepSeek-R1 for validating generalization. We employ \textbf{Qwen2.5-Math-7B-Instruct}~\citep{qwenmath} as the backbone model to perform full parameters supervised fine-tuning on selected instruction subsets. We extend the model's context length from 4,096 to 16,384 via RoPE~\citep{roformer} scaling, increasing the RoPE frequency from 10k to 300k. We conduct experiments on a Linux server equipped with 8 A100-SXM4-40GB GPUs. Utilizing the LLaMA-Factory framework~\citep{llamafactory}, we set the sequence limit of 16,384, batch size to 1, gradient accumulation steps to 4, and learning rate to 5e-5 with a warmup ratio of 0.1, followed by a cosine decay schedule towards zero. The training epochs is 3 for any size of subset. For the judge model, the rollout count is 4, and we apply the LoRA technique~\citep{lora}, with the rank of 16, alpha of 32, and dropout rate of 0.1, training for 1 epoch. We utilize DeepSpeed ZeRO-3~\citep{deepspeed} and FlashAttention2~\citep{flashattention} to accelerate computations on GPUs.

\section{Evaluations Settings}
\label{app:evaluation}
We adopt nine mathematics benchmark spanning multiple dimensions for evaluation, including three competition-level benchmarks which are AIME in 2024 \& 2025, and AMC in 2023, and six comprehensive benchmarks such as MATH-500~\citep{math500} and OlympiadBench~\citep{olympiadbench} for math reasoning, and GAOKAO in 2023 \& 2024~\citep{qwenmath}, GAOKAO MATH~\cite{qwenmath} and KAOYAN~\citep{limo} to validate the generalization capability in Chinese math. For broader reasoning tasks, we include GPQA~\citep{gpqa}, Minerva~\citep{minerva}, ZebraLogic~\citep{zebralogic} and SimpleQA~\citep{simpleqa}. Following~\citep{qwenmath, deepseek}, the system prompt for evaluation is \textit{Please reason step by step, and put your final answer within \textbackslash boxed\{\}}. For three competition-level mathematical benchmarks, 16 solutions per problem are sampled to mitigate fluctuations due to the limited number of questions, setting temperature to 0.6 and top\_p to 0.95. For benchmarks like challenging mathematics and Chinese mathematics, we employ greedy decoding by setting temperature to 0 and top\_p to 1, sampling only one solution per problem to compute accuracy. Correctness is verified using a rule-based evaluation framework~\citep{limo}, assessing numerical and formulaic accuracy. Scores are calculated referring to~\citep{qwenmath}, yielding metrics including pass rate (Pass@1) and majority voting accuracy (Maj@16). We use the vLLM inference engine~\citep{vllm} to achieve inference acceleration, where sequence is limited to 16,384.

\newpage
\section{Case study of Efficient Exploration in Reasoning}
\begin{figure}[htbp]
    \centering
    \includegraphics[width=0.78\linewidth]{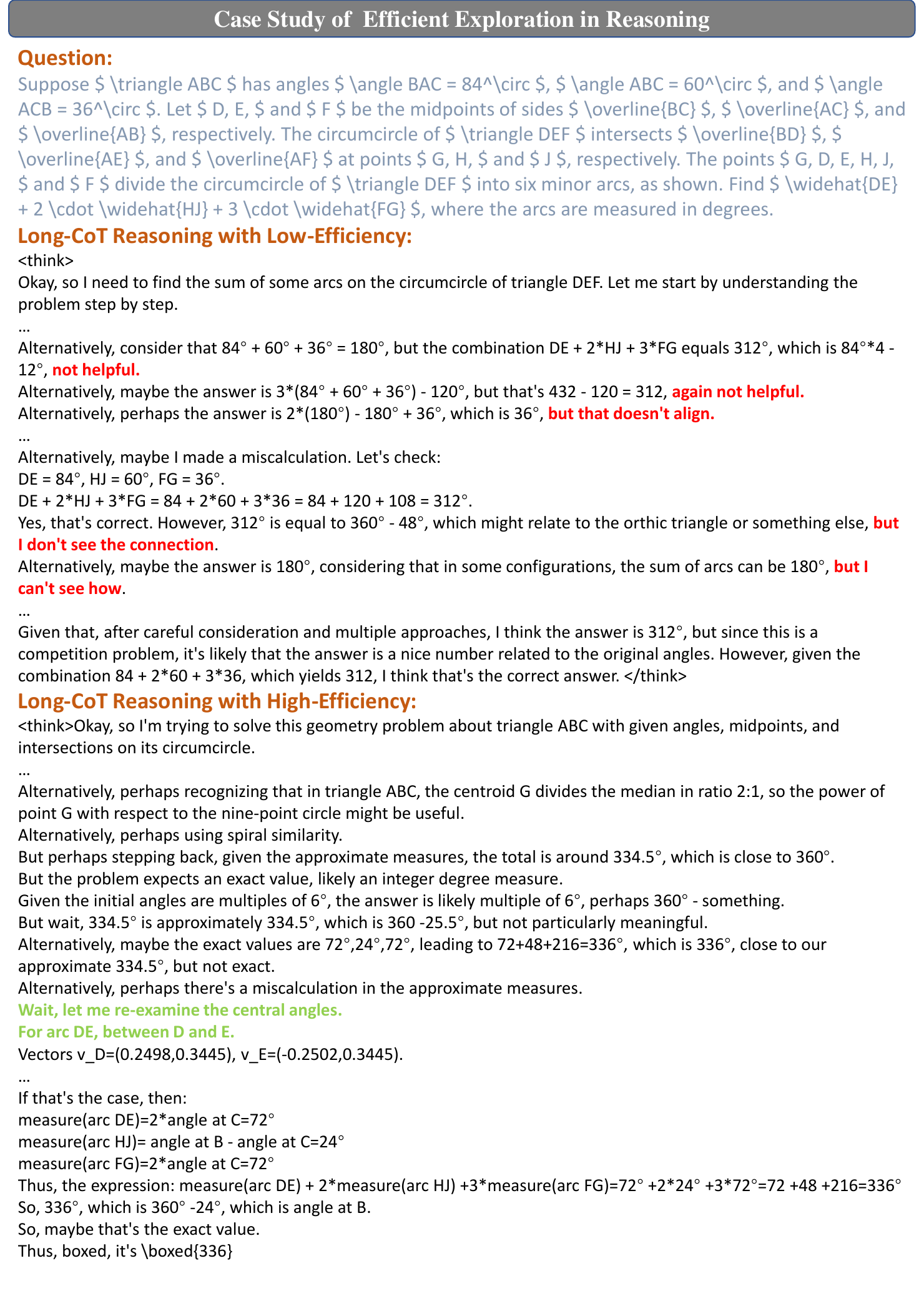}
    \caption{A case study of efficient exploration in reasoning on AIME 25. The emergence of more rethinking tokens in long-CoT reasoning does not lead to improved efficiency. In contrast, stronger LRMs reach the correct reasoning trace more efficiently.}
    \label{fig:case_efficient}
\end{figure}

\newpage
\section{Reasoning Efficiency}
\begin{figure}[htbp]
    \centering
    \includegraphics[width=0.6\linewidth]{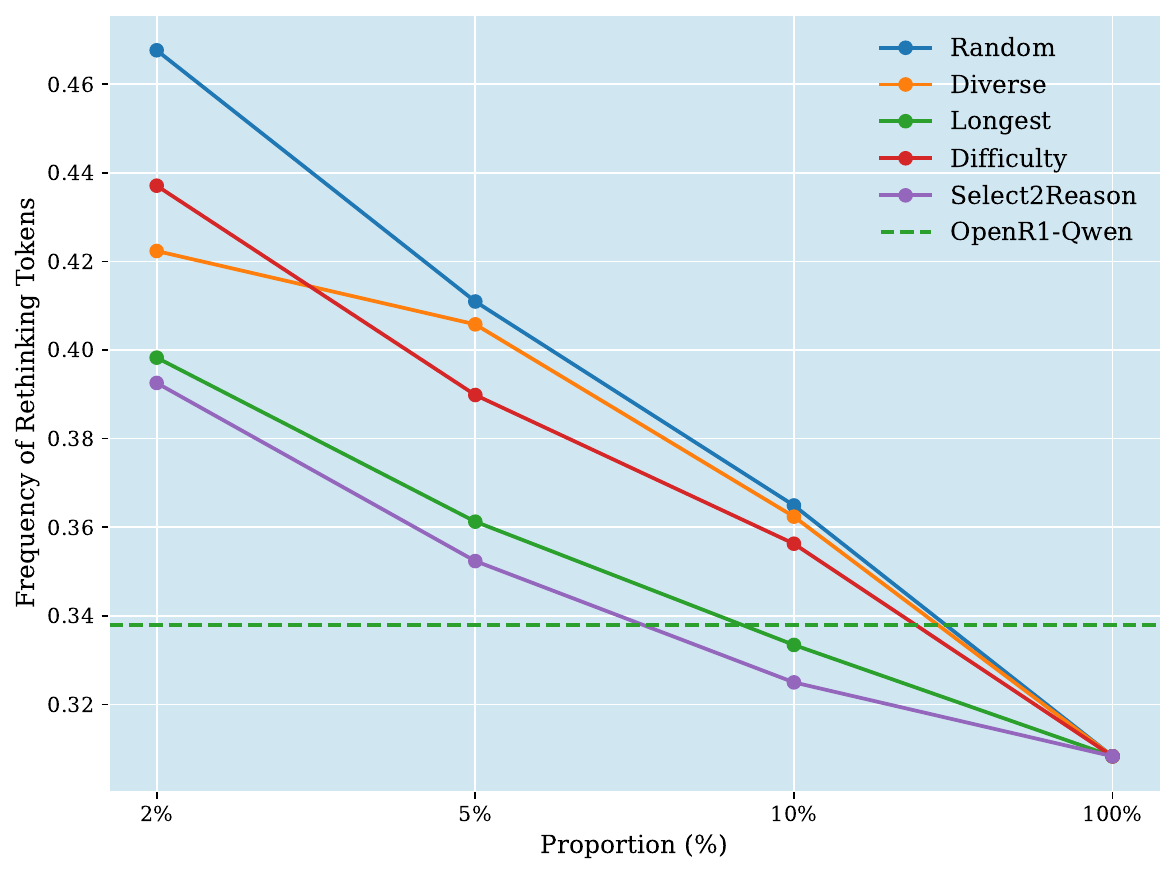}
    \caption{Relationship between performance on AIME 25 and output length.}
    \label{fig:tokens_on_aime}
\end{figure}

\section{Diversity of Instruction}
\begin{figure}[htbp]
    \centering
    \includegraphics[width=0.95\linewidth]{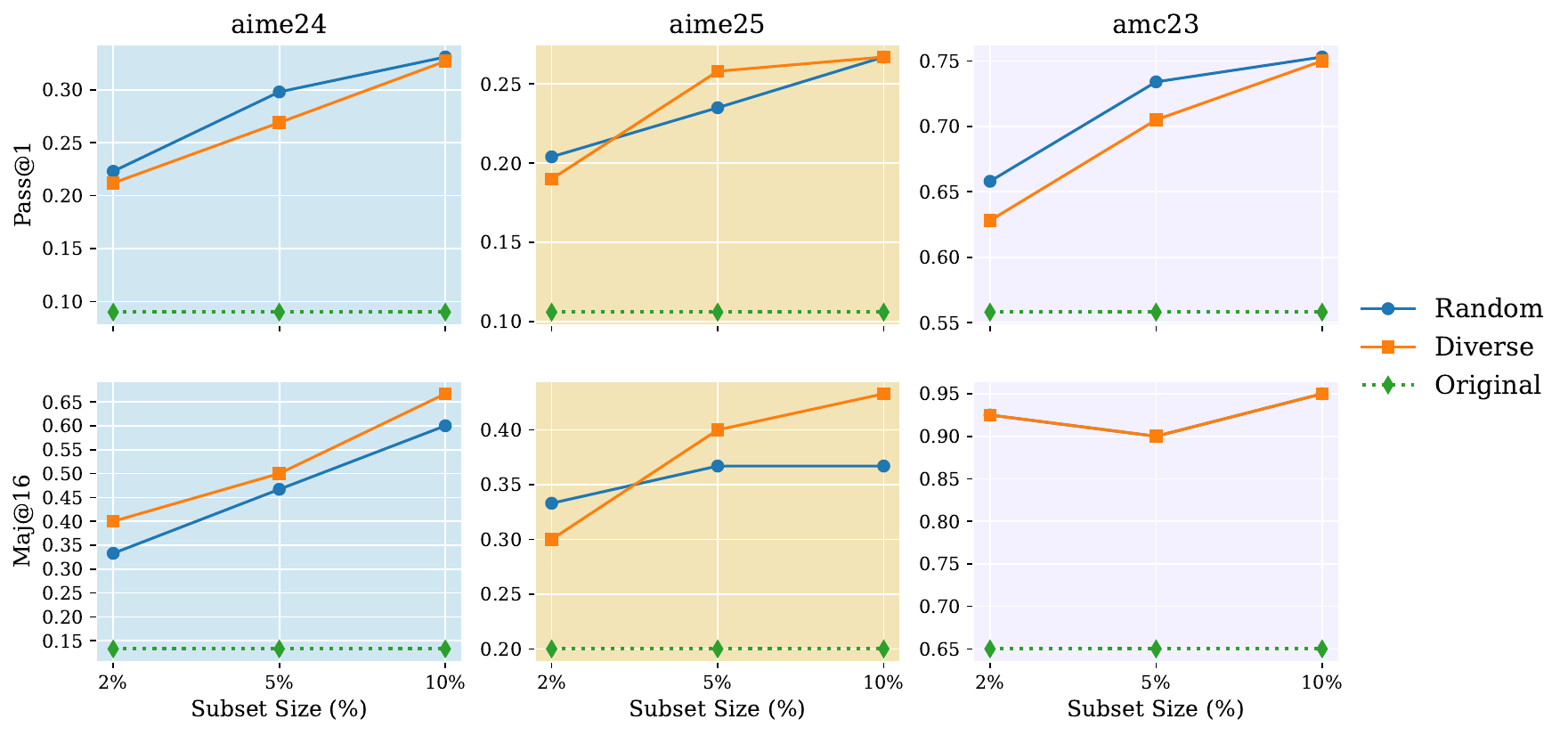}
    \caption{Performance across three expert-level benchmarks. Subset size refers to the proportion selected from data pool by length reasoning trace, either diverse or random.}
    \label{fig:diversity}
\end{figure}

\newpage
\subsection{Case study of data selection by joint ranker in Chinese-DeepSeek-R1-Distill-data.}
\label{sec:case_chinese}

\begin{figure}[htbp]
    \centering
    \includegraphics[width=1.0\linewidth]{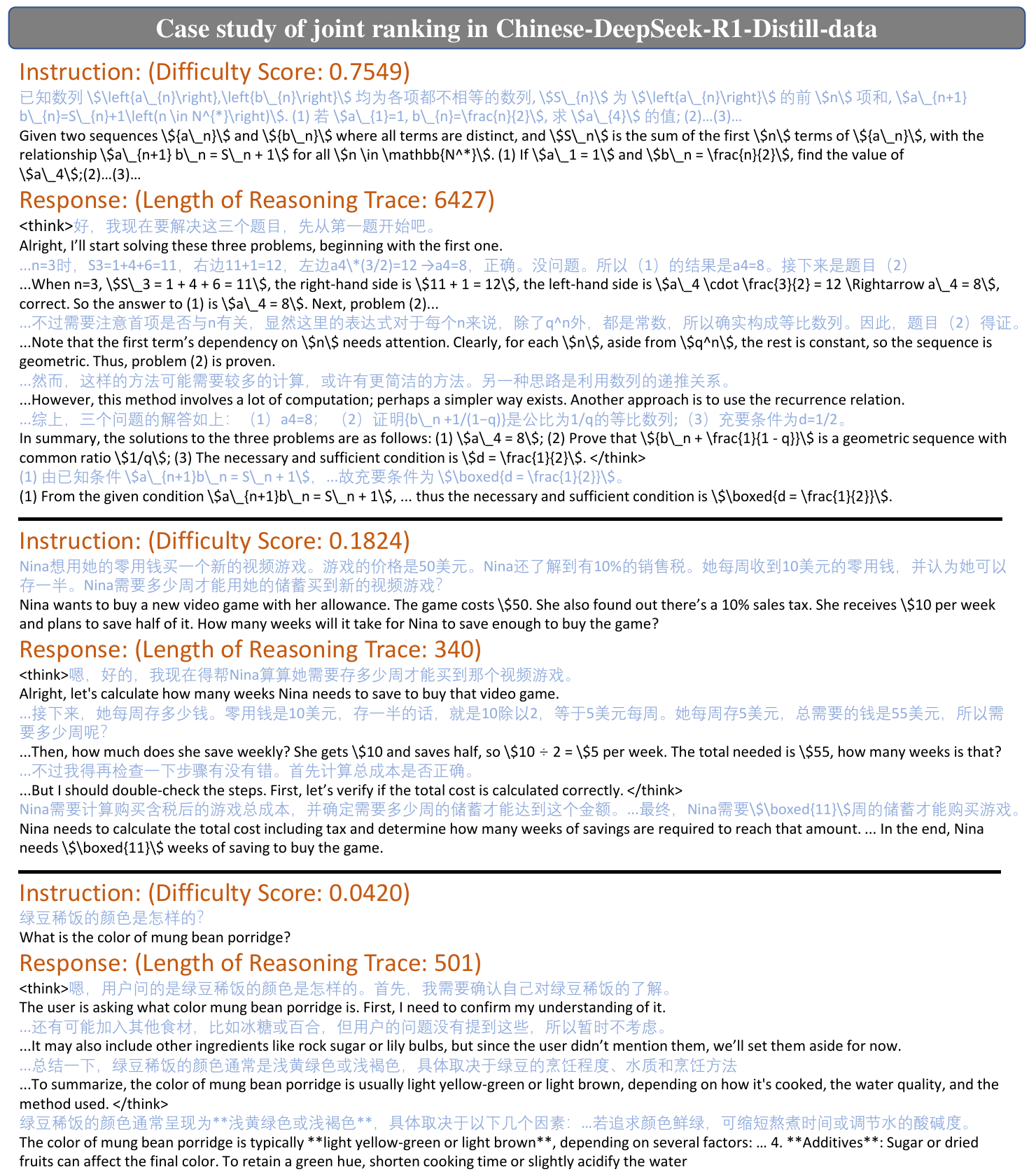}
    \caption{Case study of data selection by joint ranker in Chinese-DeepSeek-R1-Distill-data.}
    \label{fig:case_chinese}
\end{figure}

\newpage


\section{The Usage of Ai Assistants}
During the completion of this thesis, the scenarios involving the usage of Ai Assistants included: using code-completion tools to assist with experiments, and using ChatGPT to polish the draft after the initial writing was completed. Ai Assistants were not involved in any aspects such as the development of research ideas, literature review, and so on.

\end{document}